\documentclass[letterpaper, 10 pt, conference]{ieeeconf}  

\IEEEoverridecommandlockouts                              

\usepackage{graphics} 
\usepackage{epsfig} 
\usepackage{mathptmx} 
\usepackage{times} 
\usepackage{amsmath} 
\usepackage{amssymb}  
\usepackage{xcolor}
\usepackage{comment}
\usepackage{mwe}
\usepackage{caption}
\usepackage{graphicx}
\usepackage{pbox}
\usepackage{tabularx,booktabs}
\usepackage{float} 
\usepackage{subfigure}
\usepackage{comment}
\usepackage{soul}
\usepackage{gensymb} 
\usepackage[export]{adjustbox} 
\usepackage{pbox}

\usepackage{fancyhdr}
\fancypagestyle{firstpage}{%
    \chead{This paper has been accepted for publication at the 2024 International Symposium on Medical Robotics.}
    }

\title{\LARGE \bf
Towards an Autonomous Minimally Invasive Spinal Fixation Surgery Using a Concentric Tube Steerable Drilling Robot
}

\author{Susheela Sharma \IEEEmembership{Student Member, IEEE}, Sarah Go, Jeff Bonyun, Jordan P. Amadio, Mohsen Khadem,\\and Farshid Alambeigi, \IEEEmembership{Member, IEEE}
\thanks{*This work is supported by the National Institute Of Biomedical Imaging and Bioengineering of the National Institutes of Health under Award Number R21EB030796 and the Faculty Innovation Award by the University of Texas at Austin.}
\thanks{S.~Sharma, S.~Go, J.~Bonyun,  and F.~Alambeigi are with the Walker Department of Mechanical Engineering and Texas Robotics at the University of Texas at Austin, Austin, TX, 78712, USA. Email: \{sheela.sharma, sarah.go, jbonyun\}@utexas.edu,  farshid.alambeigi
@austin.utexas.edu}.
\thanks{J.~P.~ Amadio is with the Department of Neurosurgery, The University of Texas Dell Medical School, TX, 78712. }
\thanks{M.~Khadem is with the School of Informatics, University of Edinburgh, UK.}
}

\begin{document}

\maketitle
\thispagestyle{firstpage}
\pagestyle{empty}


\begin{abstract}
Towards performing a realistic  autonomous minimally invasive spinal fixation procedure, in this paper, we introduce a unique  robotic drilling system utilizing a concentric tube steerable drilling robot (CT-SDR) integrated with a seven degree-of-freedom robotic manipulator. The CT-SDR in integration with the robotic arm enables creating precise J-shape trajectories enabling access to the areas within the vertebral body that currently are not accessible utilizing existing rigid instruments. To ensure safety and accuracy of the autonomous drilling procedure, we also performed required calibration procedures. The performance of the proposed robotic system and the calibration steps were thoroughly evaluated by performing various drilling experiments on simulated Sawbone samples. 
\end{abstract}

\section{INTRODUCTION}\label{Intro} 
Continuum  manipulators (CMs) have grown increasingly popular in the last few decades because of their small size and the dexterity they can provide to surgeons \cite{BurgnerKahrs2015ContinuumRF, Sefati2021ASR}. This has allowed for advances in surgical robotics to focus on minimally invasive approaches while maintaining or exceeding the standards for an ideal outcome \cite{Vitiello2013EmergingRP,Ebrahimi2019AdaptiveCO}. However, many of the applications targeted with CMs have focused on soft tissue interventions such as \cite{Boushaki_thesis,Jiang2015DesignAA}. Only recently,  CMs targeting hard tissue surgical interventions have shown promising results and entered the field. Examples include the use of tendon-driven CMs used for treatment of pelvic osteolysis \cite{Sefati2021ASR} and femoral core decompression \cite{Alambeigi_2017}, as well as   utilizing an articulated wrist \cite{Hinged2} and concentric tube steerable drilling CM \cite{Sharma_tbme_2022,Sharma_ismr,Sharma_icra} for spinal fixation.  

Aside from proving the functionality of CMs for orthopedics and neurosurgical applications, similar to the 
commercially available systems such as  Mazor X (Medtronic, Dublin, Ireland) and the  Rosa One (Zimmer Biomet, Warsaw, IN, USA) robot assisted kit, these robots need to be integrated with another robotic arm to be properly positioned and navigated in the surgical workspace.   This integration demands utilizing a medically-approved navigation systems and calibration procedures to ensure safety, accuracy, and reliability of the robotic procedure \cite{Li2023RoboticSA}. Of note, the calibration  is the key step for performing an accurate robot-assisted or autonomous robotic  procedure. 

Despite the abundant of literature \cite{Li2023RoboticSA} focusing on developing navigation and calibration algorithms for robotic systems utilizing rigid instruments, few studies can be found on navigation and calibration of semi/autonomous robotic systems utilizing a CM for hard-tissue related interventions.  
For example, Wilkening et al. \cite{Wilkening2017DevelopmentAE} introduced basic calibration steps and control algorithms for controlling an integrated robotic system towards autonomous less invasive treatment of pelvic osteolysis using a tendon-driven CM.  Later, Sefati et. al 
 \cite{Sefati2021ASR} extended this work by performing a fully autonomous osteolytic lesion removal under optical tracking navigation and the feedback received by a sophisticated embedded  fiber Bragg Grating (FBG) sensors integrated with the utilized CM. 
Nevertheless,
 to ensure safety and success of such systems utilizing a tendon-driven CM, these robotic systems always require (1)  an active closed-loop control scheme with an embedded shape sensor  to correct for unnecessary deformation of the CM during interactions with a hard tissue; and (2) limiting the speed of cutting/drilling procedure   to avoid potential CM's structure and drilling failure while ensuring a safe procedure.
 
To address the aforementioned limitations of tendon-driven CMs, we have recently introduced a Concentric Tube Steerable Drilling robot (CT-SDR) for various orthopedics and neurosurgical interventions. In our previous publications  \cite{Sharma_tbme_2022,Sharma_ismr,Sharma_icra}, we have proved the resilience of CT-SDR in regards to spinal drilling applications. These systems have shown significantly higher inherent structural strength than the tendon driven manipulators while still being able to create straight, J- and U-shape trajectories. Despite their success, in our previous works, we solely have used a table mounted system with 1-2 degrees of freedom (DoF) \cite{Sharma_tbme_2022,Sharma_ismr,Sharma_icra}. 
\begin{figure*}[t]
	\centering
	\includegraphics[width=0.7\linewidth]{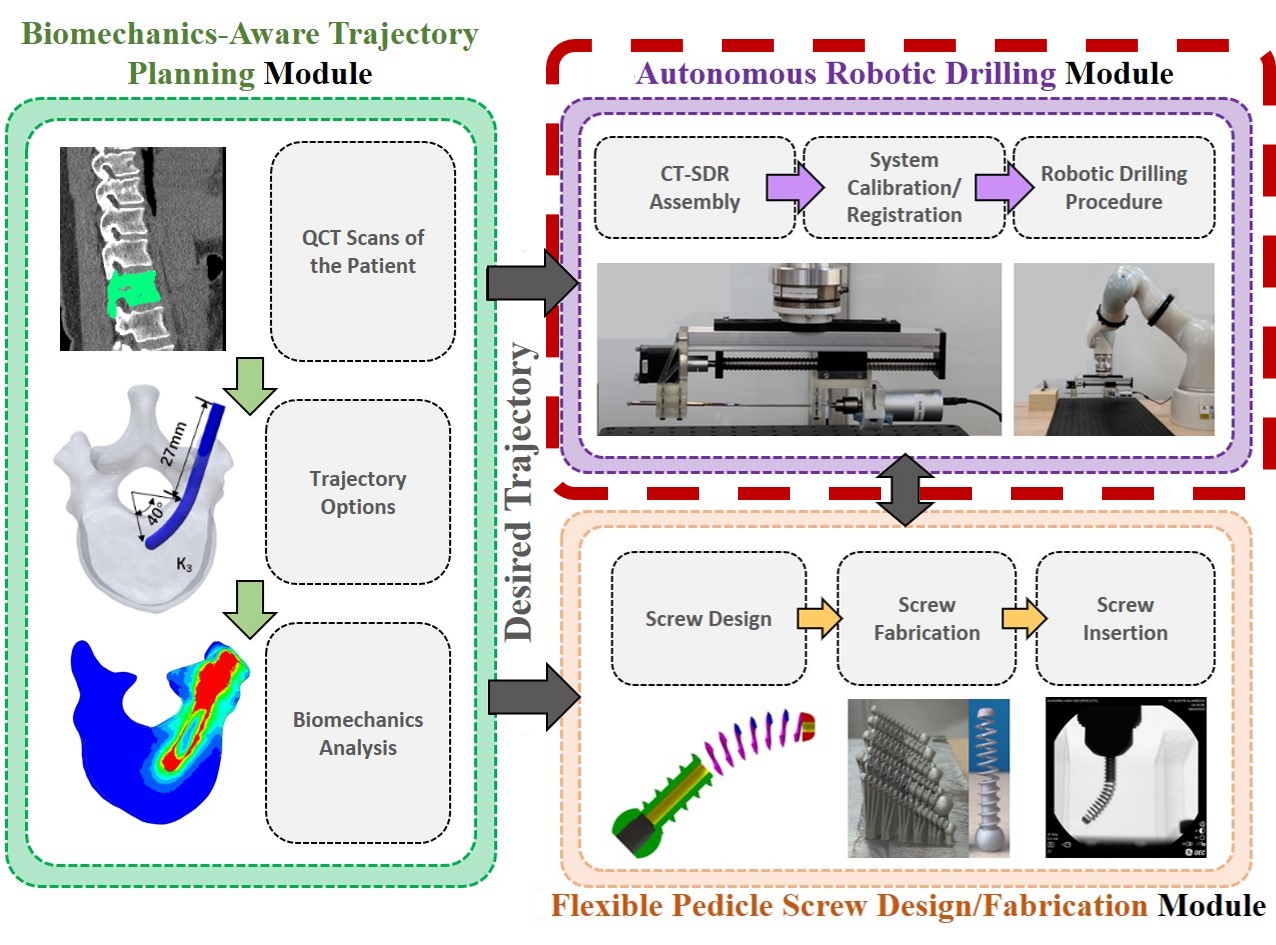}
	\caption{An outline of the overall framework for surgeon care of a patient. A \textit{Biomechanics-Aware Trajectory Planning Module} provides a desired Trajectory to both the \textit{Flexible Pedicle Screw Design/Fabrication Module} and the \textit{Autonomous Robotic Drilling Module} (highlighted as the focus of this paper). The performed fixation by these two modules is then passed to the \textit{Trajectory/Placement Analysis Module} for evaluation.}
	\label{fig:framework}
	\vspace{-8 pt}
\end{figure*}
Towards performing a realistic  autonomous minimally invasive spinal fixation surgery using our CT-SDR and as our main contributions, in this paper, we  integrate this CM with a seven degree-of-freedom (DOF) robotic arm. We  utilize the additional maneuverability granted by this arm to  position CT-SDR in a  realistic surgical workspace. To ensure safety and accuracy of the autonomous procedure, we also use a medically-approved navigation system
and perform required calibration procedures. To evaluate performance of the calibration and autonomous drilling steps, we have performed several drilling experiments in simulated bone samples.

\section{Proposed Autonomous Robotic Framework}
The framework shown in Fig. \ref{fig:framework} outlines the overall process our described autonomous  robotic drilling system will fit into. This will closely replicate what surgeons are capable of currently doing in their operating rooms while integrating new features as well. The \textit{Biomechanics-Aware Trajectory Planning Module} allows for surgeons to utilize patient scans to plan an optimal solution for their personalized fixation \cite{Sharma_ismr}. As we have shown in our previous work \cite{Sharma_ismr}, this approach allows for the selection of the most optimal trajectory to minimize the stresses and strains applied to the patient's vertebrae. 
The desired trajectory is passed to the \textit{Flexible Pedicle Screw Design/Fabrication Module} to determine ideal parameters for the selected implant \cite{Kulkarni_ismr,alambeigi2019use,bakhtiarinejad2020biomechanical}. The \textit{Autonomous Robotic Drilling Module} accepts the desired trajectory as the input which determines the trajectory settings for the CT-SDR from \cite{Sharma_tbme_2023}.
This paper will delve into the process and details within the \textit{Autonomous Robotic Drilling Module} and will assume the desired trajectory being input as the same from the results published in \cite{Sharma_ismr}. 
The following sections describe the utilized robotic framework and the calibration algorithms to accomplish an autonomous drilling procedure.

\subsection{Integrated Robotic System}
As shown in Fig. \ref{fig:set-up}, our integrated robotics system is made of three main modules. A Robotic manipulator, CT-SDR, and complementary flexible instruments.
\subsubsection{Robotic Manipulator}
The actuation unit present on the original CT-SDR shown in \cite{Sharma_tbme_2022,Sharma_ismr} was modified to integrate the robotic manipulator. 
A 3D printed coupler was made to connect the CT-SDR's base to the end effector of a KUKA LBR med 7 DoF robot arm (KUKA, Augsburg, Germany). This coupler held the CT-SDR with it's rigid outer tube in line with the KUKA's end effector's positive X-axis.

\subsubsection{CT-SDR}
The CT-SDR is assembled similarly to the design shown in \cite{Sharma_tbme_2022,Sharma_ismr}. A curved steering guide is nested within a stationary rigid stainless steel tube to create the concentric tubes implied by the name. Based on our previous study in \cite {Sharma_ismr}, for the tests performed in this paper the assumed desired trajectory from the \textit{Biomechanics-Aware Trajectory Planning Module} in Fig. \ref{fig:framework} was set at a 69.5 mm radius. Therefore a matching steering guide was used for the CT-SDR. 
During assembly, the chosen steering guide was secured so that the plane of bending was parallel to the XY plane of the robotic manipulator's end effector, and bending around the positive Z-axis. This eliminated the need for additional calibrations for the steering guide's orientation in order to map the pose of the CT-SDR's tip.

\subsubsection{Flexible Instruments}
Without loss of generality, we designed the system for work with a 9mm flexible pedicle screw similar to the implants proposed in \cite{alambeigi2018inroads,alambeigi2019use}. The CT-SDR was fitted with a flexible cutting tool with a drill bit diameter of 6 mm, allowing the system to create trajectories with an outer diameter (OD) of 7.25 mm due to vibration. Preliminary tests show that this cutting tip had little impact on the cutting capabilities of the CT-SDR. This created trajectories that align with the optimal outer diameter specifications outlined in \cite{Heidemann1998InfluenceOD}, which claim that any drilled trajectory $<85\%$ of the implants OD maximizes the implant's pullout force. 

\subsection{System Kinematics}
For the system shown in this paper, a 7 DoF robotic arm manipulator (LBR Med, Kuka) was utilized for the positioning of the attached CT-SDR as laid out in Fig. \ref{fig:set-up}. 
The forward kinematics of the  arm can be calculated using a screw theory approach, in which every joint of the robot arm is represented as an individual screw axis which applies a twist to every frame following it as you move from base to the end effector. 
We used the equation defined in (\ref{eq:screw_axis_def}) to define a screw axis for each of the seven rotational joints.
Because each joint is pure rotation, the screw pitch for each joint is equal to zero (i.e., $h=0$). The location of each axis, $q \in\mathbb{R}^3)$, and the direction of its rotation, $w\in\mathbb{R}^3$, can be found based on documentation from KUKA.
\begin{equation} \label{eq:screw_axis_def}
    S = \begin{bmatrix}w \\ v\end{bmatrix} = \begin{bmatrix}\hat{w} \\ -\hat{w}\times q +h\hat{w}\end{bmatrix}
    =\begin{bmatrix}\hat{w} \\ -\hat{w}\times q\end{bmatrix}
\end{equation}
where  $\hat{w} \in so(3)$ is the skew-symmetric matrix representation of $w$.
With the creation of each screw axis and with knowledge of the current rotation of each joint, (\ref{eq:screw_transform}) can be used to calculate the current transformation between the robot arm's base frame and end effector.
\begin{equation}\label{eq:screw_transform}
    T = [ \prod_{i=1}^{7}(e^{[S_i]\theta_i}) ]M
\end{equation}
Where $M \in SE(3)$ is the $4\times4$ transformation matrix of the end effector in the robot's base frame at the robot's home position, and $\theta$ is the rotation of the joint from its home position in radians.

As shown in  \cite{Sharma_tbme_2022}, the CT-SDR's tip is able to follow an ideal curved trajectory with a radius of 69.5 mm parallel to the XY plane of the KUKA LBR's end effector. The plane of bend is determined by the assembly of the CT-SDR as well as the initial direction. We decided to have the CT-SDR bend around the positive Z-axis and towards the negative Y-axis shown in Fig. \ref{fig:set-up} to avoid nearing joint limits on the robotic manipulator. These trajectory models are similar to the ones created in \cite{Sharma_tbme_2023}.

\subsection{Calibration Algorithms} \label{calibrate}
While many of the transformations present in the overall system, shown in Fig. \ref{fig:transforms}, are known to us through the forward kinematics of robot and also available software for   system components  (i.e. the robot arm's KUKA Sunrise and the NDI Polaris Vega Optical Tracker), there are several transformations present in the system that are unknown to us  and require performing calibration procedures to obtain. These unknown transformations are denoted by the red color in Fig. \ref{fig:transforms}.  As mentioned, to perform an accurate and safe autonomous drilling procedures, finding these   transformations are of paramount importance. The following sections describe these calibration steps.

\subsubsection{Pivot Calibration} \label{pivot}
In order to accurately determine the offset distance between the robot arm end effector and the tip of the CT-SDR integrated with the robotic arm, we performed pivot calibration. Pivot calibration is a commonly used algorithm to calculate the translational distance from a tool's tip to a manipulator's end effector in the frame of the robot's end effector. The process entailed rotating the CT-SDR around it's tip placed at a fixed point in space (referred to as the pivot). Meanwhile, we recorded different configurations of the robot arm to track the end effector's position and orientation relative to the world frame.

\begin{figure*}[t]
	\centering
	   \includegraphics [width=0.8\linewidth]{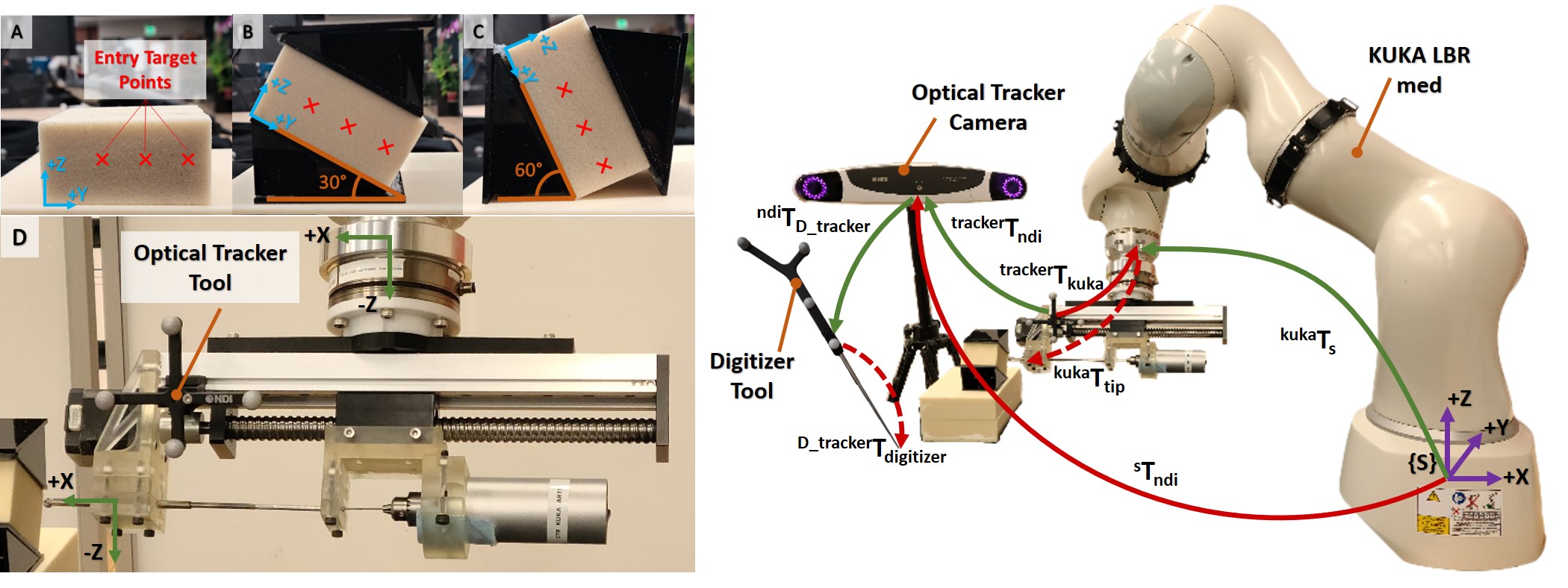}
		\caption{Experimental set-up for the CT-SDR System. Including the Optical Tracking Camera, KUKA LBR Med, and the CT-SDR(subfigure D). Marked in green are the known transforms present in our system, used for calculations of the system. Marked in red are the unknown transforms required to establish positions of any given object in space at all times. Solid dashed lines indicate pivot calibration, full lines indicate $AX=ZB$ calibration. The entry points for each test are in subfigures A-C, showing the angles that the test sample blocks were mounted at (0\degree, 30\degree, and 60\degree respectively).}
		\label{fig:set-up}
        \label{fig:transforms}
\end{figure*}

The translation is computed by establishing that the fixed pivot point's spatial location, denoted as $x_{pivot}\in\mathbb{R}^3$, can be calculated for each position of the module using the following equation  \cite{asbr}.
\begin{equation}\label{eq:pivot_pos}
x_{pivot} = R_ix_{tip}+p_i
\end{equation}
The offset estimation between the robot's end effector and the CT-SDR's drill tip, denoted as $x_{tip}\in\mathbb{R}^3$, involved the collection of several positions, each with corresponding values for the robot end effector's rotation, $R_i\in SO(3)$, and position, $p_i\in\mathbb{R}^3$. By utilizing the same stationary pivot point, both the pivot's location in the world frame and the CT-SDR's tip distance can be calculated. This is achieved by solving an overdetermined system of equations through a least-squares fit of an $Ax=b$ problem. The system takes the form:
\begin{equation}\label{eq:pivot_eq}
\begin{bmatrix} ... & ... \\ R_i & -I \\ ... & ... \end{bmatrix} \begin{bmatrix} x_{tip} \\ x_{pivot} \end{bmatrix} = \begin{bmatrix} ... \\ -p_i \\ ... \end{bmatrix}
\end{equation}
where $i$ refers to the $i$th position captured \cite{asbr}. This calculated $x_{tip}$ combined with an identity matrix $I$ creates the transformation indicated in Fig. \ref{fig:transforms} as ${}^{kuka}T_{tip}\in SE(3)$.

An identical pivot calibration process was performed on the digitizer shown in Fig. \ref{fig:transforms}, to determine the translational distance between the frame identified on the markers of the digitizer and the digitizer's tip. This translation was identified in the tool's frame and could then be applied to transforms involving the digitizer in the performed experiments. 

\subsubsection{Hand-Eye Calibration} 
The remaining missing transformations include the optical tracker in the world frame, ${}^sT_{ndi}\in SE(3)$, and the optical tracking tool on the CT-SDR in the robot arm's end effector frame, ${}^{tracker}T_{kuka}\in SE(3)$. The two missing transforms can be solved for simultaneously through a Robot-World/Hand-Eye, or $AX=ZB$, calibration process. We implemented the single step python function created by  \cite{opencv_library, Li_Wang_Wu_2010} to run the calculations of the following algorithm. The function takes a list of inputs similar to the set used in pivot calibration, where the poses of the robot arm and the corresponding optical tracker tool pose at each frame are collected by designed python script. These poses are then used to create a list of transformations of the optical tracker camera with respect to the optical tracker tool mounted on the CT-SDR, and the world frame with respect to the robot arm end effector. These transforms are made into the matrix set shown in \ref{eq:axzb}, where they take the form of an $AX=ZB$ problem.
\begin{equation}\label{eq:axzb}
\begin{bmatrix} ... \\ {}^{tracker}T_{ndi} \\ ... \end{bmatrix} 
\begin{bmatrix} X \end{bmatrix} =
\begin{bmatrix} Z \end{bmatrix} 
\begin{bmatrix} ... \\ {}^{kuka}T_{s} \\ ... \end{bmatrix}
\end{equation}
The algorithm designed by Li et al. \cite{Li_Wang_Wu_2010} and implemented into the python library cultivated in \cite{opencv_library} solves this system of equations for the two remaining transforms: $X\in SE(3)$ as the transformation from the NDI Polaris Vega (NDI, Canada) with respect to the world frame, and $Z\in SE(3)$ as the transformation from the KUKA end effector to the optical tracker tool.

With every transformation available to us, we are able to calculate a transformation matrix for every frame of interest with respect to the world frame $\{S\}$ as defined in Fig. \ref{fig:set-up}. This consistency in reference frame enables controls be commanded to the robot to move the CT-SDR's drilling tip using information gathered by the optical tracking software. An essential step in robot-assisted surgical operations.

\section{Experiments}
\subsection{Experimental Set-Up}
The instruments selected for use in our experimental set-up were included because of the existing familiarity with the operating room. Both the KUKA LBR med robotic arm and NDI Polaris Optical Tracking System are commonly used and were designed for use in the surgical field. The inclusion of these two systems to our experimental set-up ensures that the framework we implement for our procedure could be easily replicated by a surgeon in the field. Our overall set-up is shown in Fig. \ref{fig:set-up}, with the inclusion of the CT-SDR, KUKA robotic arm, and NDI Optical Tracking system. An optical tracking tool was rigidly attached to the CT-SDR for use in the calibration steps and with tracking the CT-SDR's rotation and position in space during testing. The digitizer shown in Fig. \ref{fig:transforms} was used to collect points in the world space via the NDI camera to indicate pose and position goals to the robot. The test specimen was a Sawbone bio-mechanical bone model phantom (block 5 PCF, Pacific Research Laboratories, USA). PCF 5 was selected as it simulates bone with a high level of osteoporosis \cite{ccetin2021experimental}, the target patients for the implementation of our framework.

\subsection{Experimental Procedure}
With the KUKA robot arm and attached CT-SDR at a pre-designated home position and the sawbone test sample securely mounted in front of the robot the user is prompted to provide the system with several points using the digitizer. First, the user is prompted to indicate the plane parallel with which the CT-SDR will drill its trajectory. For each test, the test specimen was held in a fixture at an angle of 0\degree, 30\degree, or 60\degree as shown in Fig. \ref{fig:set-up}A-C. These angles were arbitrarily chosen to simulate potential robot approaching angles depending on the patient's spine posture in a real surgical setting.  Experiments were performed on each angle three times for a total of 9 drilled trajectories. Through the process, a user would touch the tip of the digitizer to the top surface of the sawbone sample while saving the locations in a list. The software then takes the recorded list of 3D points to calculate a best-fit plane using a least-squares error minimization.

The user is then prompted to indicate the entry point on the test specimen for the CT-SDR with the digitizer. Using a pre-made laser cut guide held up to the face of the sample block, we ensured a reliable baseline for what we expect in our results. The guide had holes laser cut into its face to provide a location to place the digitizer to align with the sample. A goal transformation matrix is constructed from the indicated orientation and rotation and passed to the robot to align the CT-SDR. A 5 mm offset in the X-direction was included in the goal position to prevent collisions between the test sample and the CT-SDR's tip.

Upon arrival at the desired location, the CT-SDR's drilling tip was activated and accelerated to approximately 8250 rpm. The KUKA robot arm then actuated the system through a straight 13 mm (18 mm with the 5 mm offset) at a speed of 1 mm/s, using the rigid stainless steel outer tube as the primary guide tube of the CT-SDR. This allowed for the creation of a straight drilled trajectory into the sawbone block. The CT-SDR's inner curved steering guide could then be advanced through it's trajectory for 35 mm at a drilling speed of 2.5 mm/s. At the completion of the trajectory, the CT-SDR's drilling tip rotational speed was decreased to approximately 1000 rpm before retracting the curved steering guide followed by the rigid tube. The KUKA then guides the system to return to its starting home position.

\begin{figure}[t]
	\centering
	\includegraphics[width=0.65\linewidth]{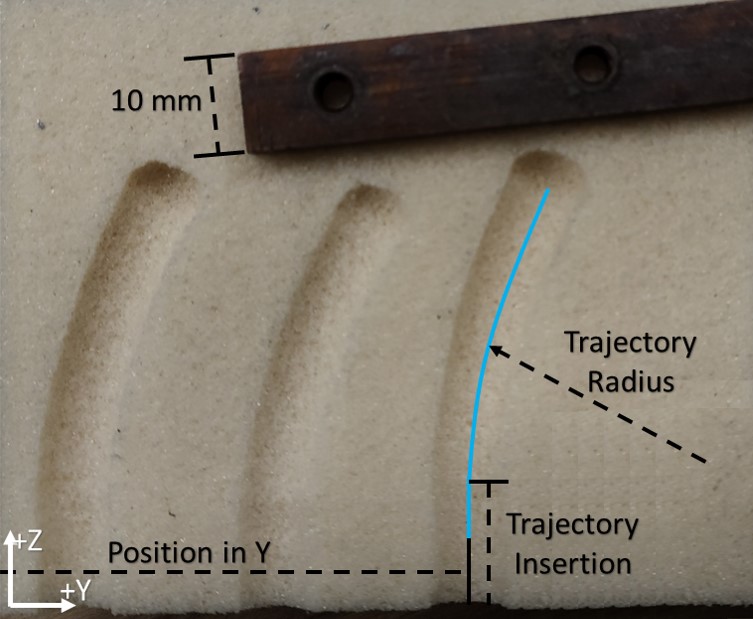}
	\caption{Cross Sectional View of several of the tests performed with the CT-SDR System with a 30\degree mounting angle.}
	\label{fig:cross_sections}
\end{figure}

\section{Results and Discussion}
The performed experiments were evaluated on three separate metrics: (1) accuracy in positional placement compared to indicated location, (2) accuracy in rotational alignment with the test specimen, (3) accuracy of the drilled trajectories to their goal.  These metrics were decided based on their relevance and importance in determining the safety of the system and it's implementation in further studies. 

\subsection{Positional Placement}

The location of the entry point into the test specimen provides a measure of accuracy for the calibration process performed when initializing the system. An offset in the optical tracking camera's position by a few millimeters can have drastic impacts on the final positioning of the CT-SDR system. The average total distance for each set of experiments was calculated and detailed in Table \ref{tab:position}. The position error was calculated from the Y and Z offset (as seen in Fig. \ref{fig:set-up}A-C) measured from each test and the difference measured in the length of the straight trajectory measured from the cut cross sections shown in Fig. \ref{fig:cross_sections}. The greatest offset in position was seen in the 60\degree rotated tests, with an error of 5.15 mm. While this is a significant offset, the standard deviation was relatively small at 0.39 mm. Standard deviation indicates the consistency of the system, and serves as a good metric for the system's repeatability. The overall average leads us to believe that the error is in the calibration process and not in the transformation calculations performed in commanding the robot's goal position. These results can also help determine future avenues to follow in improving the overall system. 

\begin{table}[t] 
	\begin{center}
	\setlength{\tabcolsep}{2pt}%
\setlength{\extrarowheight}{.9em} 
                \vspace{10pt}
			\caption{Accuracy of the CT-SDR's alignment with the digitized entry point and drilled trajectory.}
			\label{tab:position}
			\begin{tabular}{c|c|c|c|c|c}
				\hline
				\hline
				
				\pbox[c]{30ex}{\textbf{Goal} \\ \textbf{Rotation}}&\pbox[c]{30ex}{\textbf{Position} \\ \textbf{Error} \\ \textbf{[mm]}} &\pbox[c]{30ex}{\textbf{Standard} \\ \textbf{Deviation} \\ \textbf{[mm]}} &\pbox[c]{30ex}{\textbf{Average}\\ \textbf{Radius} \\ \textbf{[mm]}} &\pbox[c]{30ex}{\textbf{Standard}\\ \textbf{Deviation} \\ \textbf{[mm]}} &\pbox[c]{30ex}{\textbf{\% Error} \\ \textbf{From} \\ \textbf{Ideal}} \\
				\hline
				0\degree&3.88&0.30&70.2&2.4&1.03\\
				30\degree&4.55&0.33&70.4&1.12&1.29\\
				60\degree&5.15&0.39&72.7&3.18&4.68\\
				\hline
			\end{tabular}
	\end{center}
\end{table}

\subsection{Rotational Alignment}

\begin{table}[t] 
	\begin{center}
	\setlength{\tabcolsep}{2pt}%
\setlength{\extrarowheight}{.8em} 

			\caption{Rotational Errors on average for each set of tests.}
			\label{tab:rotation}
			\begin{tabular}{c|c|c|c|cc} 
				\hline
				\hline
				\pbox[c]{30ex}{\textbf{Goal} \\ \textbf{Rotation}}&\pbox[c]{30ex}{\textbf{Calculated} \\ \textbf{Angle}} &\pbox[c]{30ex}{\textbf{Actual} \\ \textbf{Change}} &\pbox[c]{30ex}{\textbf{Error from}\\ \textbf{Goal}} &\pbox[c]{30ex}{\textbf{Standard}\\ \textbf{Deviation}}\\
				\hline
				0\degree&1.03\degree&1.21\degree&1.21\degree&0.94\degree\\
				30\degree&29.17\degree&28.82\degree&1.18\degree&0.19\degree\\
				60\degree&62.89\degree&62.42\degree&2.41\degree&0.64\degree\\
				\hline
			\end{tabular}
	\end{center}
 \vspace{-10pt}
\end{table}

Prior to moving from the home position and then again while drilling, the pose of the optical tracker tool rigidly placed on the CT-SDR was recorded by the optical camera. From these poses, we can calculate the angle of rotation that the CT-SDR underwent while attempting to align the system with the plane of the test sample. The angle of change between the starting drilling position should be equivalent to the angle the test sample is mounted at. Because the CT-SDR started the tests at parallel to the table (the world XY plane), both the calculated command angle and the actual final drilling angle should be equal to the corresponding 0\degree, 30\degree, or 60\degree of the mounted specimen. The averaged results of the performed experiments are shown in Table \ref{tab:rotation}.

We found the results to be very accurate between the calculated goals for the robot to align with and the actual rotations felt by the CT-SDR. The smallest error across all of testing in plane alignment between the actual ending position and how the sample was mounted was 0.17\degree. This was evaluated during the 30\degree set of tests, which was the most accurate overall with an average error from the 30\degree goal of 1.18\degree with a standard deviation of 0.19\degree. The largest error was seen in the 60\degree samples where the system still maintained an average error less than 2.50\degree. 

\subsection{Trajectory Accuracy}
Measurements of the trajectory accuracy came in two parts, the accuracy of the straight trajectory and the accuracy of the curved trajectory. To measure either, the test samples were first sliced in half to fully visualize the trajectories' cross sections shown in Fig. \ref{fig:cross_sections}. Images taken of the cross sections were then analyzed using a 3D CAD Software (SolidWorks, Dassault Systèmes) to determine the straight insertion distance and the radius of curvature for the drilled trajectory.
The straight insertion distance was used to measure the accuracy in the X direction for the position of the CT-SDR's tip, while the curved trajectory assessed the accuracy of the attached tool. The maximum average error was seen in the 60\degree tests with an average error of 3.20 mm in the radius of curvature when compared to the ideal 69.5 mm defined by the desired trajectory from Fig. \ref{fig:framework}'s \textit{Biomechanics-Aware Trajectory Planning Module}. The errors seen in these experiments are very similar to other tests performed by the CT-SDR when it was table mounted in \cite{Sharma_tbme_2022}. This indicates that while mounted to a robotic arm, and additional DoF added to the system's base, the curved trajectory is unaffected when drilling through the sawbone samples.

\section{Conclusion and Future Work}

Towards implementing a robot assisted surgical system for spinal fixation surgery with flexible spinal implants, we proposed an autonomous robotic drilling system comprising of a robotic manipulator and concentric tube steerable drilling robot. The system utilized the kinematics of individual components and surgical grade instrumentation to identify crucial relationships between the required reference frames of the system. 
The maximum average error in rotation was obtained as 2.41\degree, and the drilled trajectories had a maximum radius error of 3.20 mm from the desired value. While both of these results were promising, in the future, we would like to improve on the accuracy of the system's positioning and decrease the errors seen in the system's entry position from the maximum of 5.15 mm currently measured. The system's positional alignment was shown to be very consistent with standard deviations of less than 0.4 mm in each set of experiments performed, which leads us to believe we need to improve our calibration process. Additionally, we want to integrate more realistic phantoms and evaluate our robotic system on cadaveric specimens.

\vspace{-3pt}

\bibliographystyle{IEEEtran}
\bibliography{main}

\end{document}